\pdfoutput=1

\documentclass[11pt]{article}

\usepackage{algorithm,algcompatible,amsmath}
\usepackage{amsthm,amsopn,bm}
\usepackage{algpseudocode}
\usepackage{booktabs}
\usepackage{tabularx}
\usepackage{tabulary}

\usepackage[preprint]{acl}

\usepackage{times}
\usepackage{latexsym}

\usepackage[T1]{fontenc}

\usepackage[utf8]{inputenc}

\usepackage{microtype}

\usepackage{inconsolata}

\usepackage{graphicx}

\title{RAMQA: A Unified Framework for Retrieval-Augmented Multi-Modal Question Answering}

\author{Yang Bai \and Christan Grant \and Daisy Zhe Wang \\
        University of Florida / Gainesville, Florida, USA \\
        \{baiyang94, christan, daisyw\}@ufl.edu\\}

\begin{document}
\maketitle
\begin{abstract}
Multi-modal retrieval-augmented Question Answering (MRAQA), integrating text and images, has gained significant attention in information retrieval (IR) and natural language processing (NLP). Traditional ranking methods rely on small encoder-based language models, which are incompatible with modern decoder-based generative large language models (LLMs) that have advanced various NLP tasks. To bridge this gap, we propose RAMQA, a unified framework combining learning-to-rank methods with generative permutation-enhanced ranking techniques. We first train a pointwise multi-modal ranker using LLaVA as the backbone. Then, we apply instruction tuning to train a LLaMA model for re-ranking the top-k documents using an innovative autoregressive multi-task learning approach. Our generative ranking model generates re-ranked document IDs and specific answers from document candidates in various permutations. Experiments on two MRAQA benchmarks, WebQA and MultiModalQA, show significant improvements over strong baselines, highlighting the effectiveness of our approach. Code and data are available at: \url{https://github.com/TonyBY/RAMQA}
\end{abstract}

\section{Introduction}
\begin{figure}[t]\setlength{\textfloatsep}{10pt}
    \centering
    \includegraphics[width=\linewidth]{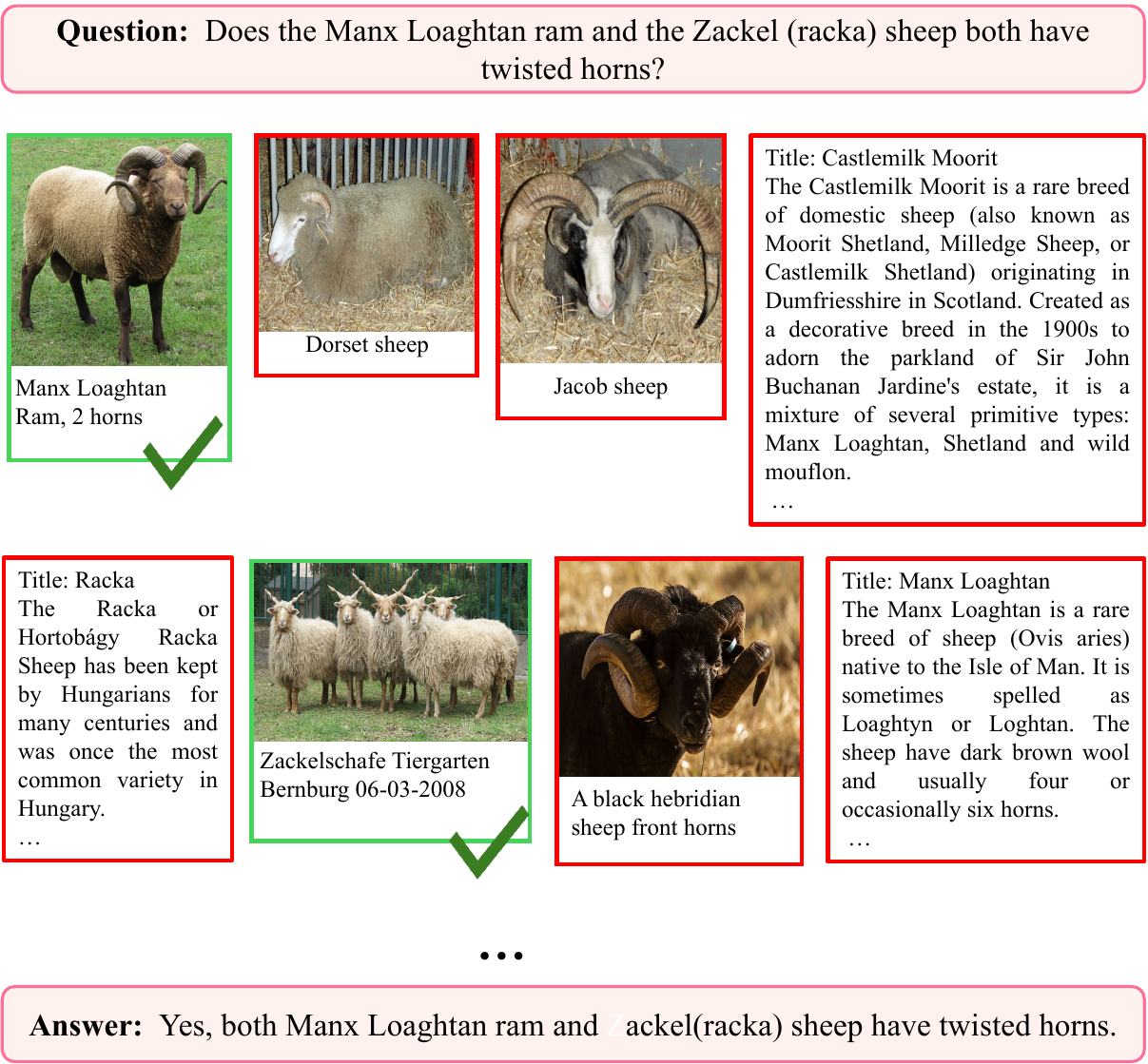}
    \caption{An example in WebQA \cite{Chang2021WebQAMA}, a Multi-modal Open-domain Question-Answering benchmark. This task requires the system to precisely identify critical sources from distractors and use these key sources to infer the answers.}
    \label{fig:WebQA-example}
\end{figure}

Multi-modal retrieval-augmented question answering (MRAQA) involves searching and integrating information from diverse modalities such as text and images \citep{MultimodalQA, Chang2021WebQAMA} (see Figure~\ref{fig:WebQA-example}). This capability is crucial for applications requiring comprehensive understanding and reasoning. While powerful generative language models have revolutionized NLP, achieving state-of-the-art results across various tasks \citep{Wu2024GPT4oVP, Touvron2023Llama2O, Liu2023VisualIT}, leveraging these advanced LLMs for information retrieval tasks like MRAQA remains challenging.

Existing MRAQA methods rely on small encoder-based ranking models \citep{51840, SKURG, Yang2023ProgressiveER}, which are not fully compatible with modern large generative language models. Although recent generative LLMs trained on massive datasets have dominated NLP tasks, they are typically decoder-only, making it challenging to encode documents into dense representations as encoder-based models do.

Generative retrieval paradigms \citep{Metzler2021, 52026, Wang2022ANC} differ from traditional retrieval methods by directly generating relevant document identifiers for a query. However, applying these methods to multi-modal information retrieval faces challenges: (1) multi-modal documents have aspects not effectively represented by static identifiers; (2) existing multi-modal LLMs are not structured or pretrained to infer across multiple multi-modal documents; (3) LLMs' limited input sequence length hinders ranking many documents in a single run.

To address these challenges, we propose RAMQA, a unified framework combining traditional learning-to-rank methods with generative ranking. First, we train a pointwise multi-modal ranker based on LLaVA \cite{Liu2023VisualIT} as a multi-modal data encoder. Second, we employ instruction tuning \cite{InstructGPT} to train a LLaMA \cite{Touvron2023Llama2O} model to re-rank the top-k documents using a novel autoregressive multi-task learning approach. Before the second-stage retrieval, we unify multi-modal documents into text representations using a zero-shot LLaVA model. This provides context for all candidate documents, reducing the LLM's burden to memorize relationships between queries and document identifiers, making it more efficient than previous methods. Our generative ranking model is trained in a multi-task manner, generating relevant documents and extracting exact answers. To reduce bias from input document sequences, we use permutations of document candidates. We demonstrate the effectiveness of these methods through comprehensive ablation studies.

Experiments on two benchmarks, WebQA \cite{Chang2021WebQAMA} and MultimodalQA \cite{MultimodalQA}, demonstrate significant improvements over strong baselines, highlighting our approach's effectiveness in enhancing multi-modal retrieval-augmented QA systems.\footnote{We achieved fourth place on the WebQA leaderboard: \url{https://eval.ai/web/challenges/challenge-page/1255/leaderboard/3168}; to our knowledge, the top three works were unpublished at submission time.}

In summary, our contributions are as follows:
\begin{itemize}
    \item {
       \textbf{Unified Framework:} We develop RAMQA, a unified framework for Retrieval-Augmented Multi-modal Question Answering, which combines traditional learning-to-rank methods with generative ranking techniques.
     }
    \item{
        \textbf{Innovative Multi-Stage Process:} We introduce a two-stage approach with a fine-tuned LLaVA for multi-modal pointwise ranking, and a fine-tuned LLaMA for generative re-ranking, enhanced by multi-task learning and document permutation techniques.

    }
    \item {
         \textbf{Comprehensive Evaluation:} We demonstrate the effectiveness of the proposed methods through a thorough ablation study and achieved significant improvements over strong baselines on two benchmark datasets, WebQA and MultimodalQA.
     }
\end{itemize}

\begin{figure*}[!tb]
    \centering
    \includegraphics[scale=0.52]{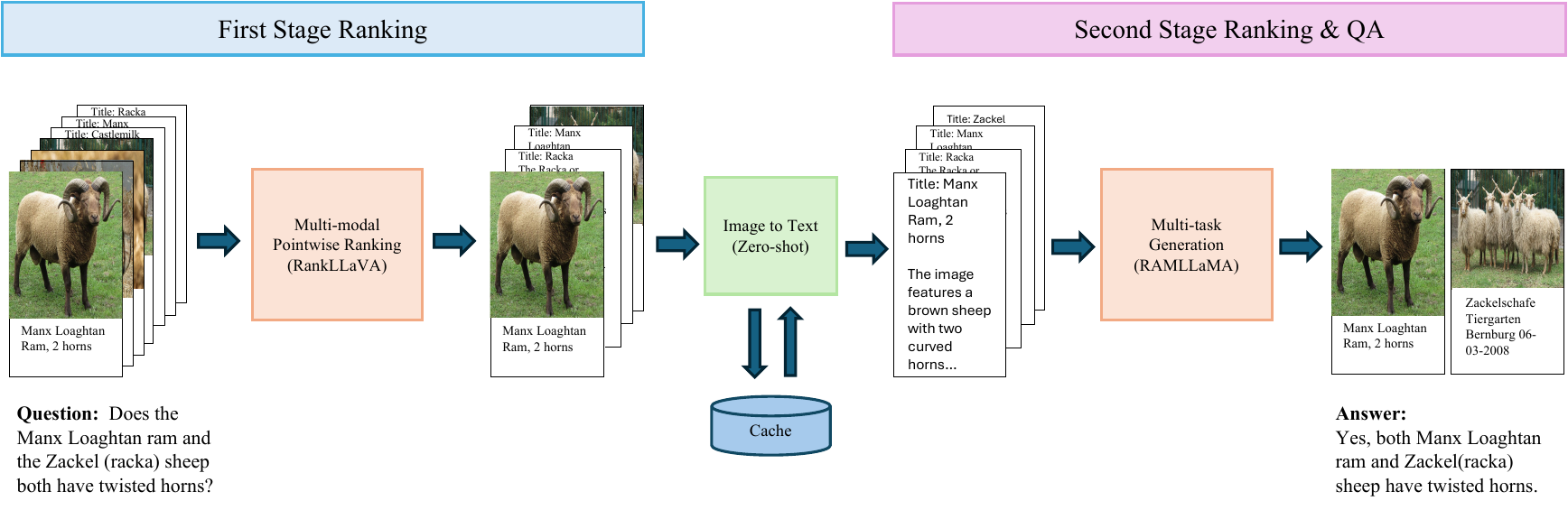}
    \caption{RAMQA Framework Overview. A detailed description of the three main components—RankLLaVA, Data Unification (Image to Text), and RAMLLaMA—is provided in Sections~\ref{sec: rankllava}, \ref{sec: data_unification}, and \ref{sec: RAMLLaMA}, respectively.} 
    \label{fig:framework_overview}
\end{figure*}

\section{Related Work}

\subsection{Multi-Modal Retrieval-Augmented Question Answering}
Multi-modal retrieval-augmented question answering (MRAQA) integrates information from various modalities, such as text, images, and tables, to answer complex questions. Benchmark datasets like MultimodalQA \cite{MultimodalQA} and WebQA \cite{Chang2021WebQAMA} have been developed to address these challenges.

Recent frameworks like MuRAG \cite{51840}, SKURG \cite{SKURG}, and PERQA \cite{Yang2023ProgressiveER} have made significant strides in MRAQA by integrating text and image data using retrieval and generation techniques. However, these methods primarily rely on encoder-based models and structured knowledge, limiting their ability to fully leverage the capabilities of state-of-the-art multi-modal generative LLMs. Our work addresses this gap by introducing a novel framework that combines traditional ranking with multi-modal generative LLMs, offering a more robust solution for MRAQA.

\subsection{Learning-to-Rank}
Learning-to-Rank (LTR) techniques optimize item ranking in information retrieval systems based on relevance. These models include pointwise \citep{10.1007/11776420_44, 10.1561/1500000016, li2011learning, Nogueira2019PassageRW, Nogueira2019MultiStageDR}, pairwise \citep{10.5555/945365.964285, DBLP:conf/iclr/ClarkLLM20, 10.1145/3600088}, and listwise \citep{10.1145/1273496.1273513, 10.1145/3341981.3344218, 10.1145/3269206.3271796} approaches. The advent of Transformer encoders like BERT \cite{Devlin2019BERTPO} and RoBERTa \cite{Liu2019RoBERTaAR} has significantly enhanced LTR by enabling more accurate relevance scoring.

Recent advancements have explored using large language models (LLMs) in LTR. For example, RankLLaMA \cite{10.1145/3626772.3657951} fine-tuned the LLaMA model, demonstrating that decoder-based LLMs can surpass traditional encoder-based models in ranking tasks. Building on this, we fine-tuned LLaVA \cite{Liu2023VisualIT}, a multi-modal LLM that combines LLaMA with the CLIP visual encoder ViT-L/14 \cite{Dosovitskiy2020AnII}, creating RankLLaVA, a multi-modal pointwise ranker that enhances ranking performance by leveraging both language and visual data.

\subsection{Generative Retrieval}
Generative retrieval techniques \citep{52026, 10.1145/3580305.3599903, NEURIPS2022_cd88d62a, 10.1145/3626772.3657797, li-etal-2023-multiview} represent a shift from traditional retrieval methods by directly generating document identifiers (DocIDs) for a query using generative models. Advances like Differentiable Search Index (DSI) \cite{52026} and SEAL \cite{NEURIPS2022_cd88d62a} have introduced more efficient and effective retrieval processes. However, these methods primarily focus on unimodal data and often struggle with integrating multi-modal information.

Our work addresses this limitation by introducing a unified framework that combines multi-modal pointwise learning-to-rank with generative ranking in a two-stage retrieval process, effectively bridging the gap in multi-modal retrieval.

\section{Methodology}
In this section, we provide a comprehensive description of our proposed framework designed to address multi-modal learning-to-rank and generative retrieval tasks. We start by defining these tasks and then explore the structure and training methodologies of our unified framework, as outlined in Figure~\ref{fig:framework_overview}.

\subsection{Preliminaries}
\subsubsection{Task Definition} 
Given a question \( Q \) and a set of input documents \( D = \{d_1, d_2, \dots, d_n\} \), where \( n \) represents the number of documents and each document may be a text with a title or an image with a caption, MRAQA aims to retrieve evidence from \( D \) and generate an answer \( A \) based on the retrieved evidence. Although the MRAQA task can encompass other document modalities like tables, audio, and video, this paper focuses specifically on text passages and images. Unlike a typical end-to-end multi-stage retrieval pipeline \cite{yates-etal-2021-pretrained}, which includes a retriever \cite{Karpukhin2020DensePR} to efficiently locate the top-k relevant texts from a corpus, followed by multiple rerankers \cite{Nogueira2019PassageRW} to refine the retrieved candidates, our approach in this paper centers on the reranking stage. Specifically, we assume the input documents include positive evidence and distractors (hard negatives) from the datasets, rather than the full document corpus.

\subsubsection{LLaMA} 
LLaMA \cite{Touvron2023Llama2O} is a large language model based on the Transformer architecture, operating in an auto-regressive, decoder-only manner. With billions of parameters, it is pre-trained on a massive dataset of web content. As a uni-directional model, its attention mechanism only considers the preceding elements in the input sequence to make predictions. Specifically, for a given input sequence \( s = [t_1, t_2, \dots, t_{n-1}] \), the model predicts the next token \( t_n \) based solely on the prior tokens. This prediction process is mathematically expressed as \( P(t_n|t_1, t_2, \dots, t_{n-1}) \), where \( P \) denotes the probability of the next token \( t_n \) in the sequence.

\subsubsection{LLaVA}  
LLaVA \cite{Liu2023VisualIT} extends the LLaMA model to handle multi-modal inputs, specifically text and images, by incorporating a vision encoder alongside its Transformer-based architecture. LLaVA retains the auto-regressive, decoder-only structure for text generation, while its vision encoder, often based on a pre-trained Vision Transformer (ViT) \cite{Dosovitskiy2020AnII}, processes images by extracting a sequence of visual features from different regions (patches) of the image.

These patch-level embeddings are then combined to form a sequence of visual tokens, which are integrated with the text tokens. The resulting multi-modal input sequence $x = [v_1, v_2, \dots, v_m, t_1, t_2, \dots, t_{n-1}]$ consists of both visual tokens $v_1, v_2, \dots, v_m$ from the image and text tokens $t_1, t_2, \dots, t_{n-1}$ from the query.

This multi-modal sequence is fed into the Transformer model, enabling it to predict the next token $t_n$ based on both visual and textual context. The prediction process is mathematically expressed as:
$P(t_n | v_1, v_2, \dots, v_m, t_1, \dots, t_{n-1})$, where $P$ represents the probability of the next token $t_n$, conditioned on both the visual features $v_1, v_2, \dots, v_m$ and prior tokens. This integration of detailed image features with text allows LLaVA to perform tasks requiring sophisticated reasoning over both visual and textual inputs, such as multi-modal question answering and image captioning.

\subsection{RankLLaVA for Multi-modal Pointwise Ranking}\label{sec: rankllava}

Our first-stage ranking model, named RankLLaVA, is trained as a pointwise ranker. This method involves feeding both the query and a candidate document into the model, which then generates a relevance score indicating how well the document matches the query \cite{Nogueira2019PassageRW}. The backbone model is initialized with LLaVA.

Traditionally, pointwise ranking models use bi-directional encoder-only models like BERT, where the [CLS] token is added at the beginning of the input sequence, and its hidden representation is used to represent the entire sequence. In contrast, since LLaVA is unidirectional, we append an end-of-sequence token ($<$/s$>$) to the input query or document, and the hidden representation of this $<$/s$>$ token is used to represent the input sequence in LLaVA.

RankLLaVA is trained on query-document pairs as detailed in Algorithm~\ref{alg:RankLLaVA}.  To compute the query-document similarity score, we utilize the LLaVA model's image encoder, tokenizer, and decoder as described in~\cite{Liu2023VisualIT}. We process the input through these components to obtain the hidden representations of the tokens. Specifically, we extract the hidden representation of the last token in the sequence from the decoder's last layer. This representation is then passed through a linear layer, and a sigmoid activation function is applied to produce the final similarity score between the query and the document.

\begin{algorithm}[t]
\small
\caption{RankLLaVA Training Procedure}
\label{alg:RankLLaVA}
\begin{algorithmic}[1]
\Require Training dataset $\mathcal{D} = \{(Q_i, d_i, y_i)\}_{i=1}^N$
\Statex \quad where $Q_i$ is a textual question, $d_i$ is a multi-modal document with image part $d_{i\_image}$ and text part $d_{i\_text}$, and $y_i$ is the ground truth label (1 if $d_i$ is relevant to $Q_i$, 0 otherwise).
\Ensure Trained RankLLaVA model

\For{each $(Q_i, d_i, y_i) \in \mathcal{D}$}
    \State \textbf{Construct Input:}
    \State Concatenate the document text with an image placeholder:
    \Statex \quad $d_i' = \text{``\texttt{<image>} } d_{i\_text} \text{''}$
    \State Construct the prompt by combining the question and the document:
    \Statex \quad $\text{prompt} = \text{``\texttt{Question:} } Q_i \text{ \texttt{Document:} } d_i' \text{ \texttt{</s>}''}$
    \State \textbf{Compute Embeddings:}
    \State Encode the image part using the image encoder:
    \Statex \quad $[v_1, v_2, \dots, v_m] = \text{ImgEncoder}(d_{i\_image})$
    \State Tokenize the prompt and obtain token embeddings:
    \Statex \quad $[t_1, t_2, \dots, t_n] = \text{Tokenizer}(\text{prompt})$
    \State Combine image features and token embeddings:
    \Statex \quad $\text{emb\_seq} = [v_1, v_2, \dots, v_m, t_1, t_2, \dots, t_n]$
    \State \textbf{Forward Pass:}
    \State Pass the combined embedding sequence through the decoder:
    \Statex \quad $\text{hidden\_states} = \text{Decoder}(\text{emb\_seq})$
    \State Extract the hidden representation of the last token:
    \Statex \quad $h_i = \text{hidden\_states}[-1]$
    \State Compute the similarity score:
    \Statex \quad $\text{Sim}(Q_i, d_i) = \sigma(\text{Linear}(h_i))$
    \State \textbf{Compute Loss:}
    \State Compute the cross-entropy loss:
    \Statex \quad $\ell_{rank\_i} = - y_i \log(\text{Sim}(Q_i, d_i)) - (1 - y_i) \log(1 - \text{Sim}(Q_i, d_i))$
\EndFor
\State \textbf{Update Model Parameters:}
\State Optimize the model parameters to minimize the total loss:
\Statex \quad $\mathcal{L} = \sum_{i=1}^N \ell_{rank\_i}$
\State \Return Trained RankLLaVA model
\end{algorithmic}
\end{algorithm}

\subsection{Multi-task Generation}
We now introduce the second stage of our framework, which functions as a multi-task generator for second-stage ranking and question answering. This stage is designed to accurately identify the correct documents from the top-k candidates predicted by the first-stage ranker that can assist in answering the question. Simultaneously, it generates the answer based on the identified documents. We experimentally show that this additional objective makes the model's ranking performance more robust.

\subsubsection{Data Unification}\label{sec: data_unification}

\begin{figure}[t]\setlength{\textfloatsep}{10pt}
    \centering
    \includegraphics[width=\linewidth]{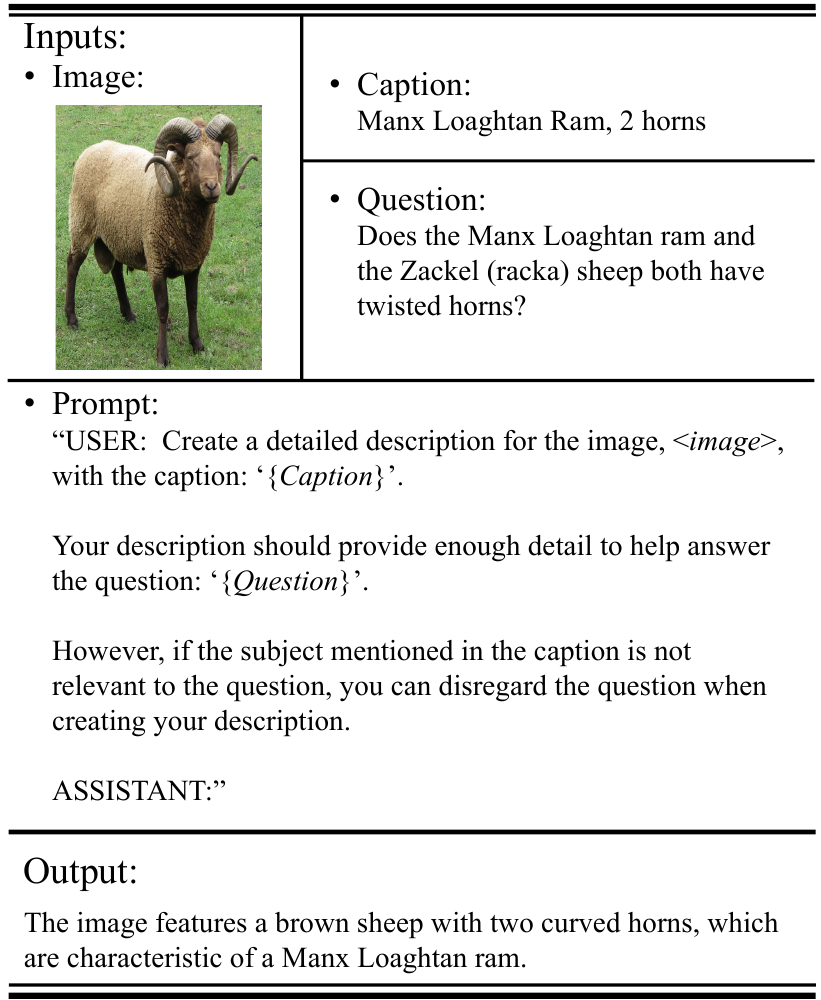}
    \caption{Zero-shot image description generation for data modality unification.}
    \label{fig:image_to_text}
\end{figure}

We begin by unifying data from different modalities by converting images to text using a pre-trained LLaVA model with a customized prompt, following the format used during LLaVA's training. Figure~\ref{fig:image_to_text} illustrates an example of our zero-shot image-to-text generation. This transformation serves two key purposes. First, it aligns with the pre-training process of LLMs, which, to our knowledge, have not been pre-trained with multiple images as input. To leverage the existing knowledge of pre-trained multi-modal LLMs and avoid costly re-training, we use LLaVA as a pointwise ranker and single-image description generator rather than as a list-wise multi-modal ranker. Second, transforming images into sentence-level descriptions optimizes input size. By conserving input token capacity, we can include more documents within the LLM's input sequence, ultimately enhancing ranking performance.

\subsubsection{RAMLLaMA}\label{sec: RAMLLaMA}
\begin{algorithm}[t]
\small
\caption{RAMLLaMA Training Procedure}
\label{alg:RAMLLaMA}
\begin{algorithmic}[1]
\Require Pre-trained LLaMA model $\mathcal{M}$, training dataset $\mathcal{D} = \{(Q_i, D_i, A_i)\}_{i=1}^N$
\Statex \quad where $Q_i$ is a question, $D_i = \{d_{i1}, d_{i2}, \dots, d_{ik}\}$ is the set of top-$k$ candidate documents, and $A_i$ is the ground-truth answer.
\Ensure Fine-tuned RAMLLaMA model $\mathcal{M}'$

\For{each $(Q_i, D_i, A_i) \in \mathcal{D}$}
    \State \textbf{Construct Input Prompt:}
    \State Randomly permute the order of $\{t_{i1}, t_{i2}, \dots, t_{ik}\}$ to get $\{t'_{i1}, t'_{i2}, \dots, t'_{ik}\}$
    \State Create input prompt $P_i$:
    \Statex \quad $P_i \leftarrow$ ``Question: $Q_i$ \textbackslash\textbackslash Documents:
    \Statex \quad \quad [DocID: 1] $t'_{i1}$
    \Statex \quad \quad [DocID: 2] $t'_{i2}$
    \Statex \quad \quad $\vdots$
    \Statex \quad \quad [DocID: $k$] $t'_{ik}$''
    \State \textbf{Construct Target Output:}
    \State Identify relevant document IDs $R_i \subseteq \{1, 2, \dots, k\}$ supporting $A_i$
    \State Create target output $T_i$:
    \Statex \quad $T_i \leftarrow$ ``Relevant Document IDs: $R_i$ \textbackslash\textbackslash Answer: $A_i$''
    \State \textbf{Fine-tune LLaMA:}
    \State Optimize $\mathcal{M}$ to minimize loss $\mathcal{L}$ over $(P_i, T_i)$:
    \Statex \quad $\mathcal{L} = -\sum_{(P_i, T_i)} \log P_{\mathcal{M}}(T_i \mid P_i)$
\EndFor
\State \Return Fine-tuned model $\mathcal{M}'$
\end{algorithmic}
\end{algorithm}

Our second-stage ranking and question-answering model, RAMLLaMA (Retrieval-Augmented Multi-task LLaMA), is trained autoregressively using instruction tuning \cite{InstructGPT}. Given a prompt comprising a question and the top-$k$ unified candidate documents from the first-stage ranking, along with their IDs, the model generates the relevant document IDs and the answer. 

To prevent the model from overfitting to the sequence of input documents, we permute the candidate documents five times for each question during training, effectively increasing the training set size fivefold. We demonstrate the effectiveness of this approach in the ablation studies.

The training procedure for RAMLLaMA is detailed in Algorithm~\ref{alg:RAMLLaMA}. Please refer to Appendix~\ref{apx: reamllama_prompt} for a training example.

\section{Experiments}
\subsection{Datasets}
We conduct experiments on two widely used MRAQA datasets: WebQA \cite{Chang2021WebQAMA} and MultimodalQA \cite{MultimodalQA}. The dataset statistics are presented in Table~\ref{tab:benchmark_stats}.

\subsubsection{WebQA} \label{sec: webQA}
WebQA \cite{Chang2021WebQAMA} contains multi-hop, multi-modal question-answer pairs, where each query, typically requiring 1-2 images or text documents, is paired with around 40 multi-modal distractors (hard negatives). Although the input sources are multi-modal, the questions are entirely text-based. Answers are free-form sentences. Evaluation metrics include source retrieval F1 and a QA score, which combines BARTScore-based \cite{yuan2021bartscore} fluency and relevance (QA-FL) with keyword accuracy (QA-Acc). The overall QA score, a product of QA-FL and QA-Acc, is the key metric for WebQA.

\subsubsection{MultimodalQA}
MultimodalQA \cite{MultimodalQA} contains multi-modal QA pairs across tables, texts, and images, with 16 question types, 13 of which require cross-modal retrieval and reasoning. As tables are outside the scope of our paper, following \cite{51840} we focus on the subset of queries involving only text and image information, specifically selecting questions labeled as ‘TextQ’ or ‘ImageQ’. Each query typically requires 1 image and/or 1 text snippet to answer and is paired with around 20 visual and text distractors. Since test set labels are unavailable, we report RAMQA results on the validation set. The answers are spans or short phrases, and the evaluation metrics are Exact Match (EM) and average F1 as described in \cite{Dua2019DROPAR}. 

\begin{table}[tb]
\centering
\resizebox{1.0\linewidth}{!}{
\begin{tabular}{lccc}
\toprule
Dataset      & Train     & Dev       & Test      \\
       & Image/Text & Image/Text & Image/Text \\
\midrule
WebQA        & 18K/17K   & 2.5K/2.4K & 3.4K/4K   \\
MultimodalQA & 3.6K/7.5K & 371/721   & -       \\
\bottomrule
\end{tabular}
}
\caption{Overall Statistics of benchmark datasets.}
\label{tab:benchmark_stats}
\end{table}

\subsection{Baselines}
We compare RAMQA against SOTA models\footnote{Note, neither MuRAG and PERQA have published their code.} on WebQA and MultimodalQA in an distractor setting, i.e., the input documents are positives and hard negatives provided by the datasets, rather than the entire document corpus.

\subsubsection{AutoRouting}
AutoRouting \cite{MultimodalQA} converts multi-modal QA into unimodal QA by using a question-type classifier to identify the modality likely to contain the final answer. It directs the question and input sources to the appropriate QA module (textQ, tableQ, or imageQ) and extracts answer spans using specialized sub-models. AutoRouting employs RoBERTa-large \cite{Liu2019RoBERTaAR} for question-type classification as well as textQ and tableQ, while VILBERT-MT \cite{Lu201912in1MV} handles imageQ with image features extracted by Faster R-CNN \cite{Ren2015FasterRT}.

\subsubsection{VLP and VLP + VinVL}
Leveraging VinVL \cite{Zhang2021VinVLRV} for image feature extraction, these transformer-based encoder-decoder models begin by concatenating each document with the question and employing a classifier to estimate the selection probability of each document. The selected documents, along with the question, are then concatenated and fed into the model for answer generation, using a beam search with a size of 5.

\subsubsection{MuRAG} 
MuRAG \cite{51840} is pre-trained on a combination of large-scale image-text and text-only corpora. It retrieves the Top-K nearest neighbors from a memory of image-text pairs using a query \( Q \) from any modality. The retrieved results are combined with \( Q \) and fed into an encoder-decoder for answer generation. During fine-tuning, the question is used as the query \( Q \) along with the Top-4 retrieved sources, and a beam search with size 2 is applied. MuRAG uses ViT-large \cite{Dosovitskiy2020AnII} for image encoding and T5-base \cite{Raffel2019ExploringTL} for text encoding and answer generation. MuRAG is evaluated only on the text and image subsets of MultimodalQA, excluding the table modality.

\subsubsection{SKURG}
SKURG \cite{SKURG} integrates evidence features using entity relations and feeds them into a transformer to generate key evidence and answers. It employs OFA-base \cite{Wang2022OFAUA} as the image encoder and BART-base \cite{Lewis2019BARTDS} for text and knowledge graph encoding. The BART decoder is then used to generate the relevant document IDs and answers from the encoded documents. The BART-base model is pre-trained on SQuAD2.0 \cite{squard2}. For entity and relation extraction, SKURG uses ELMo-based \cite{ELMo} NER \cite{NER} and OpenNRE \cite{OpenNRE}, respectively.

\subsubsection{PERQA}\label{sec: PERQA}
PERQA \cite{Yang2023ProgressiveER} is a framework for evidence retrieval and question answering. After preprocessing all images by generating descriptions using image captioning with OFA \cite{Wang2022OFAUA} and object detection with Fast RCNN \cite{fast_rcnn}, it performs iterative pairwise ranking using BERT \cite{Devlin2019BERTPO}, followed by an extra "evidence refinement" using pointwise reranking with Deberta-large \cite{deberta}. Once the top candidate documents are retrieved, PERQA integrates them into a dialogue format and fine-tunes a multi-modal LLM mPLUG-Owl \cite{mPLUG-Owl}, to generate answers based on the retrieved documents and the question.

\subsection{Implementation Details}
The backbone of RankLLaVA is based on the LLaVA-1.5-7B model\footnote{\url{https://huggingface.co/llava-hf/llava-1.5-7b-hf}} \cite{Liu_2024_CVPR}. We added a linear layer to project the final layer’s end-of-sequence token representation into a scalar, as detailed in section~\ref{sec: rankllava}. Parameter-efficient fine-tuning (PEFT) techniques, including Quantization \cite{Jacob2017QuantizationAT} and low-rank adaptation (LoRA) \cite{hu2022lora}, were used to fine-tune the model on a single NVIDIA A100 80GB GPU with a maximum input sequence length of 2048, a batch size of 8, and gradient accumulation steps of 4. With LoRA, only the linear layer parameters of the LLM were updated, while all other layers, including the visual encoder, were kept frozen.

The backbone of RAMLLaVA is based on the LLaMA-3-70B model\footnote{\url{https://huggingface.co/meta-llama/Meta-Llama-3-70B}} \cite{dubey2024llama3herdmodels}. Following the instruction tuning approach outlined in Section~\ref{sec: RAMLLaMA}, we fine-tuned the model using similar PEFT methods. This enabled fine-tuning on a single NVIDIA A100 80GB GPU with a maximum input sequence length of 8192 tokens. We employed a batch size of 2 with 16 gradient accumulation steps. The input data comprised the top 15 ranked documents.

\begin{table}[!tb]
\centering
\resizebox{1.0\linewidth}{!}{
\begin{tabular}{lccccc}
\toprule
        Model               & QA-FL      & QA-Acc    & QA & Retr-F$_1$              \\ 
        
        \midrule
        VLP(Q-only)~\citeyearpar{Chang2021WebQAMA} & 34.9 &22.2 &13.4  & --    \\
        
        VLP~\citeyearpar{Chang2021WebQAMA} & 42.6 &36.7 &22.6     &68.9          \\
        
        VLP + VinVL~\citeyearpar{Chang2021WebQAMA} & 44.2 &38.9 &24.1    &70.9    \\
        
        MuRAG~\citeyearpar{51840} & 55.7 &54.6 &36.1 &74.6   \\ 
        
        SKURG~\citeyearpar{SKURG} & 55.4   & 57.1   & 37.7    & 88.2 \\

        PERQA~\citeyearpar{Yang2023ProgressiveER} & \underline{61.7}   & \underline{63.9}   & \underline{44.4}   & \textbf{89.6} \\

        \midrule
        RAMQA (\textbf{ours}) & \textbf{64.1}   & \textbf{66.6}   & \textbf{48.1}  & \underline{88.4}\\
        \bottomrule 
\end{tabular}
}
\caption{WebQA official test set results indicated on leaderboard\protect\footnotemark~as of August 2024.  VLP (Q-only) uses only the question as input for VLP. Bold numbers indicate best and underline the second-best score.}

\label{webqa_result}
\end{table}

\footnotetext{\url{https://eval.ai/web/challenges/challenge-page/1255/leaderboard/3168}}

\begin{table}[!tb]
\centering
\resizebox{1.0\linewidth}{!}{
\begin{tabular}{lccccccc}
\toprule
    &\multicolumn{2}{c}{Text}  &\multicolumn{2}{c}{Image}   &\multicolumn{1}{c}{All}\\
    Model & EM        &F$_1$      & EM        &F$_1$     & EM \\ 
    
    \midrule
                      
    Q-only~\citeyearpar{MultimodalQA}         &  15.4  & 18.4  & 11.0  & 15.6  &  13.8  \\
        
    AutoRouting~\citeyearpar{MultimodalQA}      &  49.5  & 56.9  & 37.8  & 37.8  &  46.6  \\

    MuRAG~\citeyearpar{51840}                 & 60.8   & 67.5  & \underline{58.2}  & 58.2  &  60.2 \\
  
    SKURG~\citeyearpar{SKURG}                 & 66.7   & 72.7  & 56.1  & 56.1  &  \underline{64.2}  \\ 

    PERQA~\citeyearpar{Yang2023ProgressiveER} & \underline{69.7}   & \underline{74.1}  & 54.7  & \underline{60.3}  &  62.8  \\

    \midrule
    
    RAMQA (\textbf{ours}) &  \textbf{79.5}  & \textbf{85.5} & \textbf{67.0} &  \textbf{67.0}  & \textbf{70.6} \\

    \bottomrule
\end{tabular}
}
\caption{MultimodalQA dev-set results on the subset. Q-only denotes using only the question as input for BART- large. Bold numbers indicate best and underline the second-best score. }
\label{tab:multimodalqa_result}
\end{table}

\subsection{Main Results}
We compare RAMQA against the most relevant methods, including SOTA models.

Table~\ref{webqa_result} presents the results on WebQA. For the QA score, which is the most critical metric in the WebQA benchmark (described in Section~\ref{sec: webQA}), RAMQA outperforms all baselines, exceeding the SOTA PERQA by 8.3\% overall. In terms of Fluency, RAMQA surpasses PERQA by 3.9\%, and in Accuracy, it outperforms PERQA by 4.2\%. These improvements highlight the high fluency and accuracy of RAMQA’s generated answers. In retrieval performance, RAMQA is on par with the SOTA model PERQA. However, unlike PERQA, which relies on textual information retrieval after extensive image processing (generating captions and extracting objects) as described in Section~\ref{sec: PERQA}, RAMQA employs true multi-modal IR, directly extracting ranking features from images in the first-stage ranking.

The MultimodalQA results are presented in Table~\ref{tab:multimodalqa_result}. RAMQA significantly outperforms all baselines. For text questions, our model achieves a 14.0\% improvement in Exact Match (EM) over the SOTA PERQA. For image questions, the gap is even more pronounced, with a 15.1\% improvement over the SOTA MuRAG. Overall, RAMQA surpasses the second-best PERQA by 9.9\% in EM.

\begin{table}[!tb]
\centering
    \resizebox{1.0\linewidth}{!}{
        \begin{tabular}{l|cccc}
        \toprule
        Model & QA-FL & QA-Acc & QA & Retr-F$_1$ \\
        
        \midrule
        
        RAMQA                    & \textbf{63.4}$\pm$0.7   & \textbf{66.2}$\pm$0.4   & \textbf{47.5}$\pm$0.6   & \textbf{88.3}$\pm$0.1  \\

        w/o Perm                 & 62.4$\pm$0.9   & 64.3$\pm$0.6   & 46.2$\pm$0.7   & 86.2$\pm$0.2  \\

        Retr-only Gen w/o Perm   & -      & -      & -      & 84.7$\pm$0.2  \\

        QA-only Gen              & 58.6$\pm$1.1   & 60.8$\pm$0.8   & 40.4$\pm$0.9   & 75.4$\pm$0.1  \\ 

        \bottomrule
    \end{tabular}
}
\caption{Ablation study of RAMQA on the WebQA test set. "Perm" refers to generative retrieval with permutation. "Retr-only Gen" and "QA-only Gen" indicate generation with only the retrieval objective and with only the question-answering objective, respectively. The best results for each metric are highlighted in bold. The results are averaged over three runs with different random seeds.}
\label{tab:multimodal_ablation}
\end{table}

\subsection{Ablation Studies}
\subsubsection{Effectiveness of Permutation-Based Generative Retrieval and Multi-Task Objective Generation.}
In this section, we investigate the impact of Permutation-based Generative Retrieval and the multi-task objective generation on the final MRAQA results over the WebQA test set. 
As shown in Table~\ref{tab:multimodal_ablation}, without the generative retrieval objective, our second-stage generation model achieves an overall QA score of only 40.4. The retrieval F$_1$ score here reflects the ranking performance of our first-stage model, RankLLaVA. Documents are selected if their binary classification confidence exceeds a specified threshold, determined through tuning on the WebQA development set.

When we introduce the retrieval generation objective during the fine-tuning of our second-stage generative model, both the QA score and retrieval F$_1$ score see significant improvements. Specifically, the QA score increases by 14.4\%, and the retrieval F$_1$ score rises by 14.3\%. This demonstrates the effectiveness of the multi-task objective generation in enhancing the model's generative capabilities.

Furthermore, by introducing the permutation of candidate documents in the training data, the retrieval F$_1$ score is boosted by an additional 2.6\%, and the QA score improves by 4.1\%. This indicates that permutation-based generative retrieval not only enhances the model's retrieval performance but also contributes to a better understanding of context, thereby improving overall QA performance.

\begin{figure}[!tb]\setlength{\textfloatsep}{10pt}
    \centering
    \includegraphics[width=\linewidth]{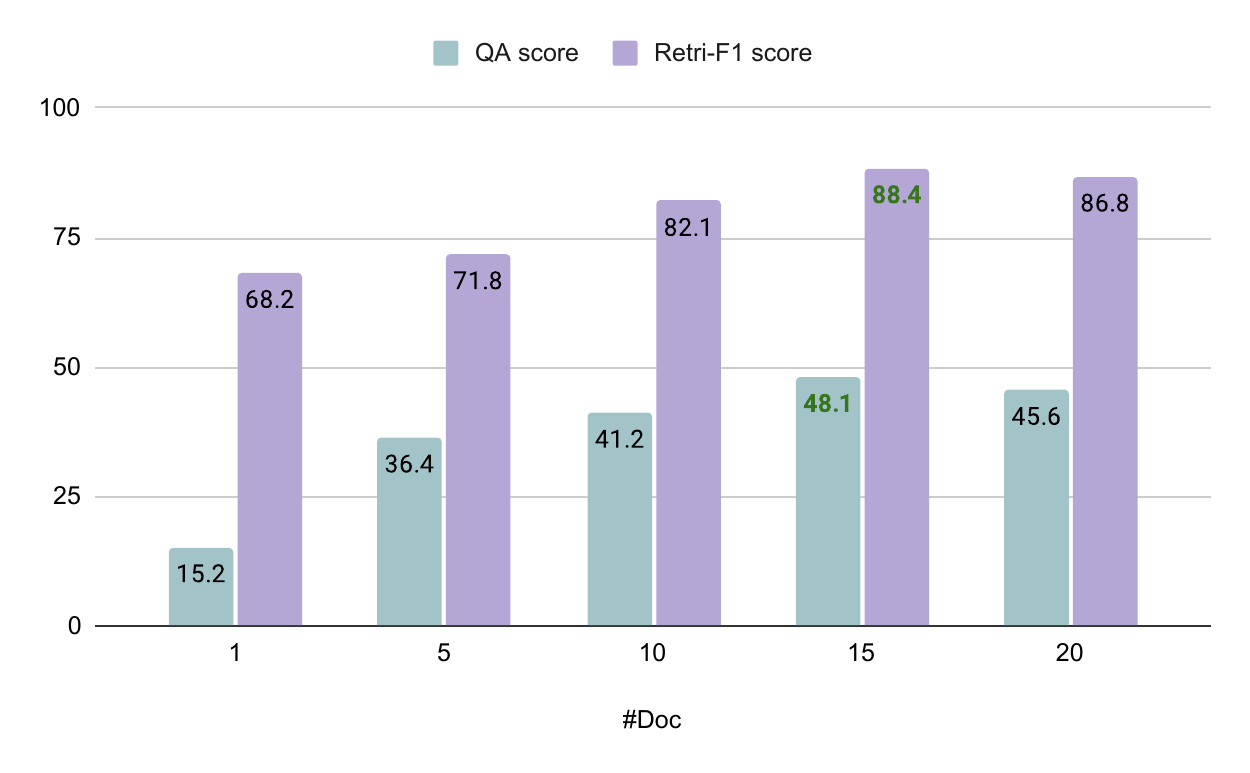}
    \caption{Impact of Input Length on RAMLLaMA Performance on the WebQA test set. The horizontal axis(\#Doc) represents the number of candidate documents from RankLLaVA's output included in RAMLLaMA's input prompt during both training and testing. We ensured that the input prompt length did not exceed LLaMA3's limit of 8096 tokens in any of the experiments.}
    \label{fig:ablation_input_length}
\end{figure}

\subsubsection{Impact of Document Count on Ranking Effectiveness.}
We investigated the impact of the number of input documents on the performance of our second-stage generation model, RAMLLaMA. Figure~\ref{fig:ablation_input_length} illustrates how varying the number of documents in RAMLLaMA’s input affects its final performance. We found that increasing the input document count from 1 to 15 improved retrieval performance, suggesting that the model benefited from the higher recall provided by the larger document set. However, increasing the input to 20 documents resulted in a performance decline. This drop is likely due to the lack of additional recall from the top 20 retrieved documents compared to the top 15, combined with the inclusion of less relevant documents, which made it more challenging for RAMLLaMA to process the input effectively, potentially leading to overfitting on irrelevant details.

\section{Conclusion}
In this paper, we introduced RAMQA, a unified framework for Retrieval-Augmented Multi-modal Question Answering that combines traditional learning-to-rank methods with generative permutation-enhanced ranking techniques to address the challenges of multi-modal retrieval-augmented question answering. By leveraging state-of-the-art generative LLMs like LLaVA and LLaMA, RAMQA significantly improves both retrieval accuracy and question-answering performance across diverse data sources, including text and images.

Experiments on two MRAQA benchmarks, WebQA and MultiModalQA, demonstrate significant improvements compared to strong baselines, highlighting the effectiveness of our approach in enhancing multi-modal retrieval-augmented QA systems. The introduction of permutation-based generative retrieval and multi-task learning objectives played a key role in these advancements, contributing to a better understanding of context and more accurate information retrieval.

In conclusion, RAMQA sets a new benchmark in multi-modal question answering, demonstrating the effectiveness of combining traditional and generative approaches. We anticipate that RAMQA and similar models will continue to advance the capabilities of multi-modal information retrieval and generation.

\section*{Limitations}
While RAMQA demonstrates strong performance and introduces several innovative techniques in multi-modal question answering, it is important to acknowledge its limitations: (1) Dependency on High-Quality Multi-Modal Data: RAMQA’s performance is closely tied to the quality and diversity of the multi-modal data available during training. In scenarios where such data is scarce or noisy, the model's ability to accurately retrieve and generate relevant answers may degrade. This limitation is particularly evident in domains where multi-modal datasets are limited or not well-structured. (2) Generalization to Novel Domains: While RAMQA has demonstrated strong results on the WebQA and MultimodalQA datasets, its ability to generalize to entirely new domains or query types remains uncertain. The model may struggle with domain-specific terminology or data formats that were not encountered during training, limiting its applicability in specialized fields. (3) Bias and Ethical Concerns: Despite its sophisticated design, RAMQA is not immune to biases present in the training data. These biases can be reflected in the retrieval and generation processes, leading to outputs that may reinforce existing stereotypes or omit crucial perspectives. Addressing these ethical concerns requires further research and careful consideration.

By recognizing these limitations, we hope to guide future research efforts aimed at overcoming these challenges and improving the robustness, scalability, and ethical integrity of multi-modal question answering systems like RAMQA.

\section*{Acknowledgments}
Our sincere thanks go out to the anonymous reviewers who took the time to provide constructive comments. 

\bibliography{main}

\appendix

\newpage

\section{Training example for RAMLLaMA}\label{apx: reamllama_prompt}
For this case study, we qualitatively evaluated the model's capabilities by sampling examples from the WebQA development dataset. We compared the model's outputs against the benchmark's golden answer set and strong baseline models.

\begin{figure}[th]\setlength{\textfloatsep}{10pt}
    \centering
    \includegraphics[width=\linewidth]{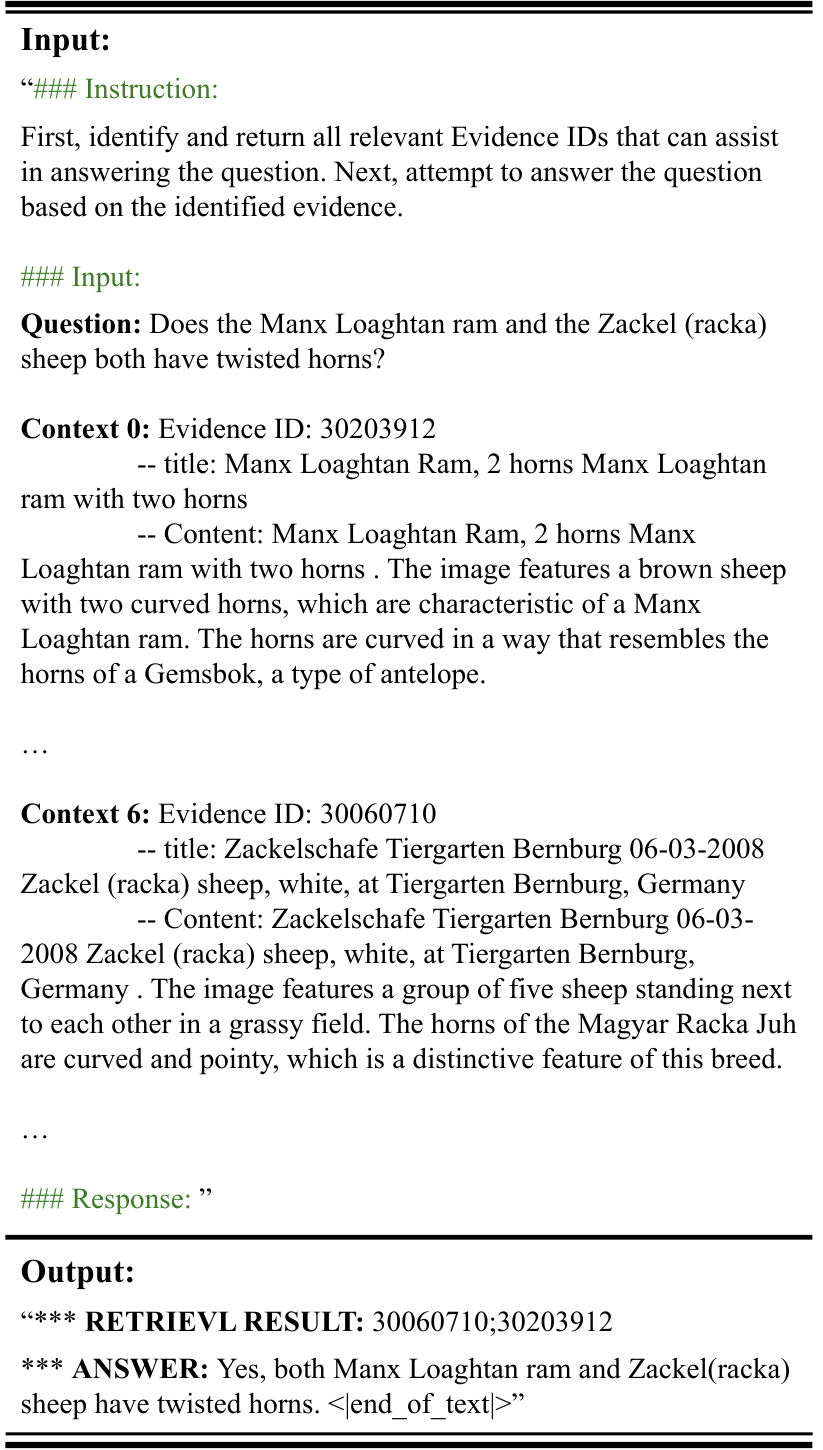}
    \caption{A Training data example of RAMLLaMA.}
    \label{fig:ramllama_prompt}
\end{figure}

Figure~\ref{fig:ramllama_prompt} illustrates a training example for RAMLLaMA, formatted following Stanford-Alpaca\cite{alpaca}.

\section{Case Study}
\begin{figure*}[!tb]
    \centering
    \includegraphics[scale=0.48]{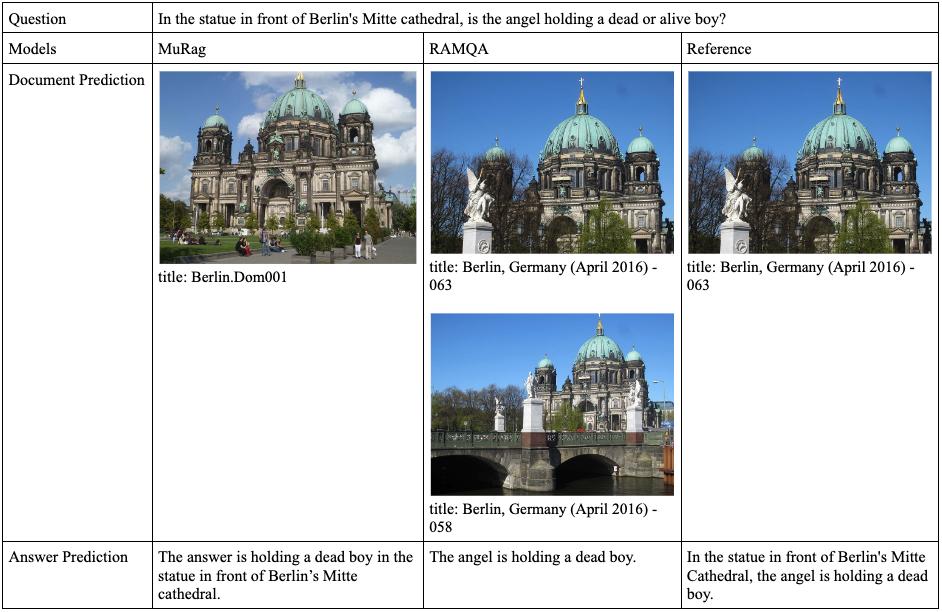}
    \caption{Prediction Examples of RAMQA vs. MuRAG} 
    \label{fig:case_study_1}
\end{figure*}

Figure~\ref{fig:case_study_1} compares the outputs of RAMQA (our model) and MuRAG QA as reported in their paper. RAMQA correctly identified the document annotated in the golden set, which MuRAG missed, and also predicted an additional document. Although this second document isn't in the golden set, it should be considered correct, as it contains crucial information. Even though it occupies only a small portion of the image, it provides the necessary details to answer the question.

\begin{figure*}[!tb]
    \centering
    \includegraphics[scale=0.57]{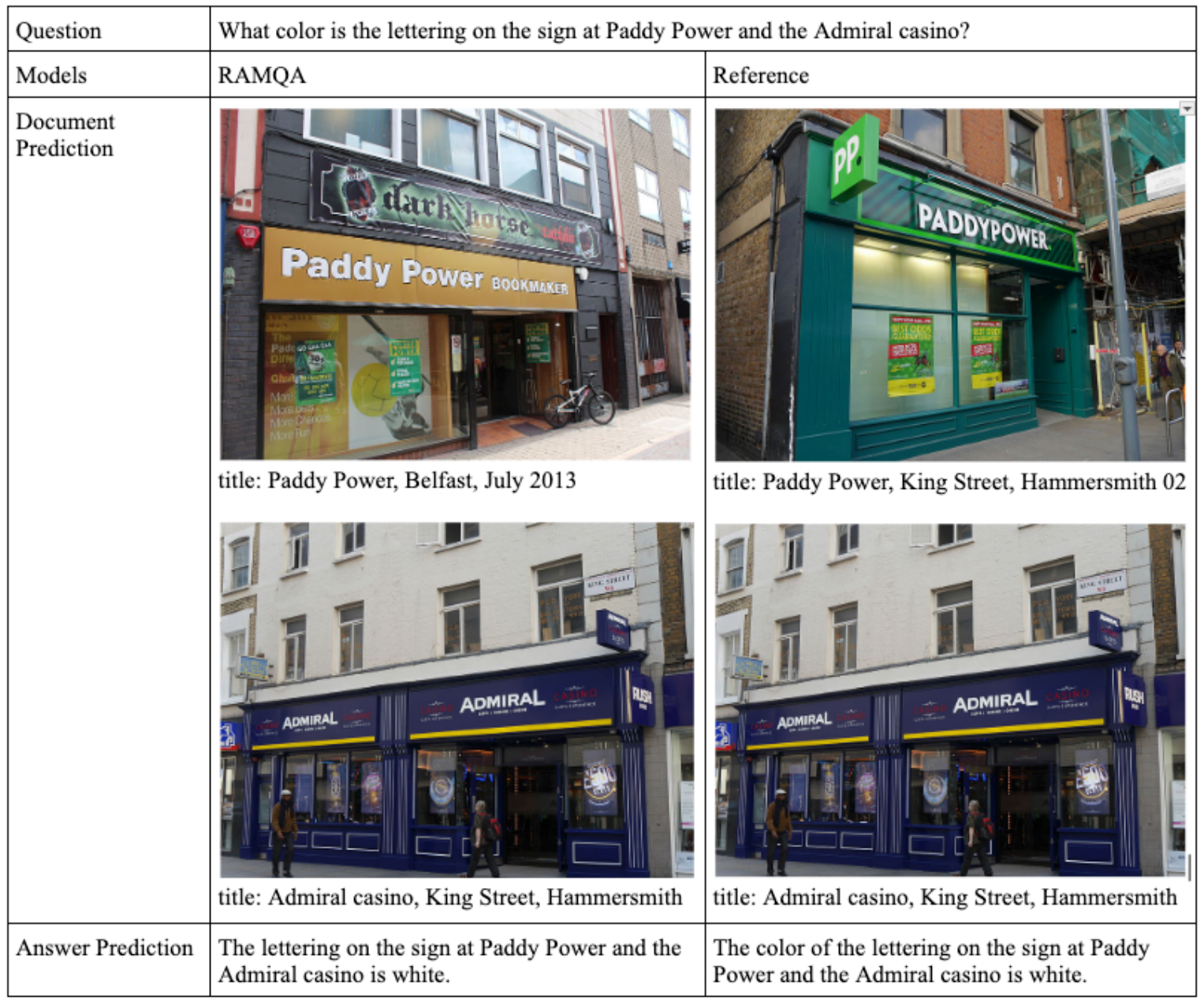}
    \caption{An Example of RAMQA Mispredictions Caused by Incomplete Document Annotations.}
    \label{fig:case_study_2}
\end{figure*}

Figure~\ref{fig:case_study_2} presents an instance where RAMQA made a misprediction. Although the answer was absent from the benchmark's golden set, it was factually correct. This highlights a limitation in the benchmark dataset, where the golden set may not fully encompass all valid answers. As a result, strict reliance on standard evaluation metrics may undervalue the model's true performance. Future work should consider expanding the golden answer set and employing more flexible evaluation methods, such as human judgment, for a more accurate assessment.

\begin{figure*}[!t]
    \centering
    \includegraphics[scale=0.56]{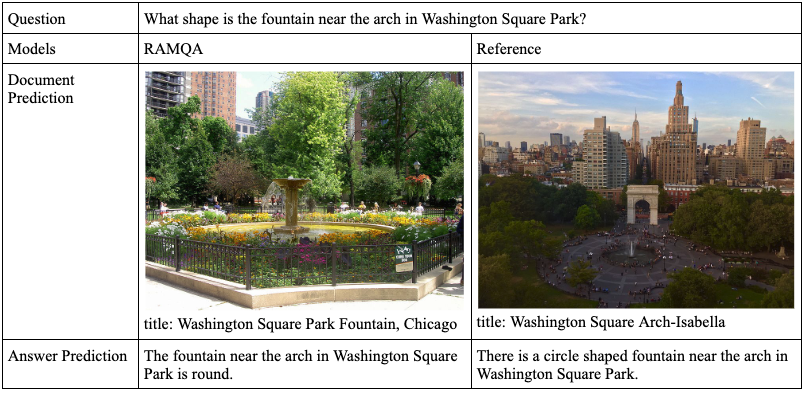}
    \caption{An Example of RAMQA Misprediction Due to Inability to Distinguish Similarly Named Locations} 
    \label{fig:case_study_3}
\end{figure*}

Figure~\ref{fig:case_study_3} shows another example of RAMQA's misprediction. This error likely stems from the model's difficulty in distinguishing between similarly named locations and their landmarks. For example, both New York City and Chicago have parks named Washington Square with fountains, but only New York’s park features an iconic arch. The model may have focused on the shared name rather than the unique characteristics of each location. This confusion could be due to misaligned image retrieval, a lack of contextual understanding, or overlapping training data. To prevent such errors, the model should prioritize distinctive features, like New York’s arch, when processing queries involving similar entities.

\newpage

\section{Scientific Artifacts}
The licenses for the resources used in this paper are as follows: MultiModalQA (MIT License), WebQA (CC0-1.0 License), LLaVA (Llama 2 Community License), LLaMA (Llama 3 Community License Agreement), and Huggingface Transformers (Apache License 2.0). We have adhered to the intended use of all referenced artifacts in this paper.

\end{document}